\makeatletter
\newcommand{\newlineauthors}{%
  \end{@IEEEauthorhalign}\hfill\mbox{}\par
  \mbox{}\hfill\begin{@IEEEauthorhalign}
}
\makeatother

\documentclass[10pt,conference,review]{IEEEtran} 
\IEEEoverridecommandlockouts

\usepackage{cite}
\usepackage{amsmath,amssymb,amsfonts}
\usepackage{algorithmic}
\usepackage{graphicx}
\usepackage{textcomp}
\usepackage{xcolor}
\usepackage{comment}
\usepackage{balance}
\usepackage{xspace}
\usepackage{booktabs}
\usepackage{tabularray}
\usepackage[ruled, lined, linesnumbered, commentsnumbered, longend]{algorithm2e}
\usepackage{multirow,multicol}
\usepackage{tikz}
\usepackage{xcolor}
\usepackage{booktabs}
\usepackage{tabularx}
\usepackage{colortbl}
\usepackage{array}
\usepackage{rotating}
\usepackage{enumitem}
\usepackage{tcolorbox}
\usepackage{subfigure}
\usepackage{fontawesome}

\usepackage[hidelinks]{hyperref}
\hypersetup{
    colorlinks=false,
    linkcolor=black,
    filecolor=black,      
    urlcolor=black,
    }
\usetikzlibrary{matrix,chains,positioning,decorations.pathreplacing,arrows}
\definecolor{mygreen}{RGB}{0, 128, 0}   
\definecolor{myorange}{RGB}{255, 165, 0} 
\definecolor{myteal}{RGB}{0, 128, 128}   

\SetCommentSty{mycommfont}

\newcommand{\ourTool}{PromptExp\xspace}

\newcommand{\boxc}[1]{\begin{tcolorbox}[left=1pt,right=1pt,top=0pt,bottom=0pt,colback=gray!5,before skip=5pt,after skip=5pt]#1\end{tcolorbox}}

\def\BibTeX{{\rm B\kern-.05em{\sc i\kern-.025em b}\kern-.08em
    T\kern-.1667em\lower.7ex\hbox{E}\kern-.125emX}}
\usepackage[utf8]{inputenc}
\DeclareUnicodeCharacter{FF0C}{,}
    
\begin{document}


\title{PromptExp: Multi-granularity Prompt Explanation of Large Language Models}

\author{
  \textbf{Ximing Dong\textsuperscript{1}},
  \textbf{Shaowei Wang\textsuperscript{2}},
   \textbf{Dayi Lin\textsuperscript{1}},
  \textbf{Gopi Krishnan Rajbahadur\textsuperscript{1}},\\
  \textbf{Boquan Zhou\textsuperscript{3}},
  \textbf{Shichao Liu\textsuperscript{3}},
   
  \textbf{Ahmed E. Hassan \textsuperscript{4}} 
  \\
  \textsuperscript{1}Huawei, Canada
   \textsuperscript{3}Huawei, China,
  \textsuperscript{2}University of Manitoba,
  \textsuperscript{4}Queen's University
\\
\small{
    \{ximing.dong,dayi.lin, gopi.krishnan.rajbahadur1,liushichao2,zhouboquan\}@huawei.com， 
 shaowei.wang@umanitoba.ca, ahmed@cs.queensu.ca
  }
  
}

\maketitle
\begin{abstract}
Large Language Models (LLMs) excel in tasks like natural language understanding and text generation. Prompt engineering plays a critical role in leveraging LLM effectively. However, LLM’s black-box nature hinders its interpretability and effective prompting engineering. A wide range of model explanation approaches have been developed for deep learning models (e.g., feature attribution-based and attention-based techniques) However, these local explanations are designed for single-output tasks like classification and regression, and cannot be directly applied to LLMs, which generate sequences of tokens. Recent efforts in LLM explanation focus on natural language explanations, but they are prone to hallucinations and inaccuracies.
To address this, we introduce \ourTool, a framework for multi-granularity prompt explanations by aggregating token-level insights. \ourTool introduces two token-level explanation approaches: (1) an aggregation-based approach combining local explanation techniques (e.g., Integrated Gradient), and (2) a perturbation-based approach with novel techniques to evaluate token masking impact. \ourTool supports both white-box and black-box explanations and extends explanations to higher granularity levels (e.g., words, sentences, components), enabling flexible analysis. We evaluate \ourTool in case studies such as sentiment analysis, showing the perturbation-based approach performs best using semantic similarity to assess perturbation impact. Furthermore, we conducted a user study to confirm \ourTool's accuracy and practical value, and demonstrate its potential to enhance LLM interpretability.
\end{abstract}

\section{Introduction}\label{sec:introduction}
Large Language Models (LLMs) are rapidly proving their potential across a multitude of tasks, consistently achieving state-of-the-art performance in areas such as natural language understanding, text generation, and more~\cite{min2023recent}. However, due to the black-box nature of LLMs, practitioners lack effective approaches to understanding the behaviors of LLMs and assist them in prompt engineering (e.g., prompt debugging and prompt refinement) and model improvement~\cite{ashtari2023discovery}. 

Due to the ``blackbox'' nature of the deep learning (DL) models and the importance of model explanation, enormous of model explanation approaches has been developed for deep learning models~\cite{shickel2020sequential,zhao2024explainability} For instance, one family of techniques called feature attribution-based explanation, which is aim to measure measure the relevance of each input feature (e.g., words, phrases, text spans) to a model’s prediction~\cite{zhao2024explainability}, such as perturbation-based explanation~\cite{wu2020perturbed,li2015visualizing}, Gradient-based explanation~\cite{hechtlinger2016interpretation,sundararajan2017axiomatic,sikdar2021integrated}, surrogate models~\cite{ribeiro2016should,lundberg2017unified}, and decomposition-based methods~\cite{du2019attribution}. Another family is called attention-based explanation, in which attention mechanism is often viewed as a way to attend to the most relevant part of inputs~\cite{zhao2024explainability,jaunet2021visqa}. 
However, the majority of those local explanation techniques, which are typically designed for tasks that produce a \textit{single} output (e.g., classification and regression). However, the output for LLMs is a sequence of tokens involving \textit{multiple} output tokens, and the majority of the existing local explanation techniques cannot be applied directly. A recent direction for LLM explanation is natural language explanation, in which the explanation is directly generated by the LLM~\cite{wei2022chain,camburu2018snli}. However, natural language explanations remain susceptible to hallucination or inaccuracies~\cite{chen2023models,ye2022unreliability}. 

To address this, we introduce \ourTool, a framework that assigns importance scores to each component of a prompt at multiple levels of granularity by aggregating token-level explanations. To implement token-level explanation, we propose two approaches: 1) aggregation-based approach, which aggregates results from existing local explanation techniques (e.g., Integrated Gradients (IG)~\cite{hechtlinger2016interpretation}) across time steps to form a comprehensive explanation for each token in the prompt. 2) perturbation-based approach, in which we adapt the existing perturbation-based approach by introducing three techniques to assess the impact of masking tokens, depending on available constraints (e.g., whether output logits are provided). Our token-level explanation covers both the while-box and black-box explanation, so that practitioners can choose based on their working context. 
Furthermore, \ourTool provides the flexibility to extend token-level explanations to higher levels of granularity, such as words, sentences, or components, by summing the importance scores of relevant tokens. This flexibility helps practitioners understand and refine complex prompt components like few-shot examples or instructions, aligning with a developer’s mental model.

To evaluate the effectiveness of \ourTool, we conduct a case study on sentiment analysis and to evaluate the Flip rate after perturbing the most important tokens generated by \ourTool, on GPT-3.5 and Llama-2. 
Our results show that the perturbation-based approach using semantic similarity to measure the impact of perturbation performs the best among all proposed approaches. 
Furthermore, we conduct a user study to evaluate the quality of the explanation generated by \ourTool and its usefulness in practice. The results show that in more than 80\% of the cases, participants found the explanation generated by \ourTool is reasonable and accurate. Participants also found that the explanations generated by \ourTool are useful in assisting them with tasks in practice, such as understanding the model's behavior of misclassification cases and compressing prompts for saving inference time and cost. 

We summarize our contribution as follows:

\begin{itemize}
    \item We propose \ourTool, a framework for prompt explanation for LLMs, which first provides the feasibility for Multi-granularity explanation.
    \item We developed various token-level explanation approaches and conducted an extensive evaluation.
    \item We offer practical insights on selecting suitable token-level explanation techniques within \ourTool.
\end{itemize}



\section{Methodology}\label{sec:method}

Before introducing our framework, we first define our problem. Given a prompt $P$ = \{$p_1,p_2, ..., p_n$\} and an LLM $L$, the goal of our framework is to provide the importance score of each token in the prompt, $imp_i$, which reflects their influence on upcoming output by LLM.

To enable this, we propose \ourTool, a framework to explain prompts in multi-granularity by aggregating the token-level explanation. To achieve the token-level explanation, two approaches are implemented: 1) aggregation-based approaches and 2) perturbation-based approaches. 
We also offer the flexibility to extend prompt explanation to multiple levels of granularity, such as word, sentence, or component by summing up the importance scores of corresponding tokens. We elaborate on two approaches for token-level explanation and the multi-granularity explanation below.

\subsection{Aggregation-based Token-level Explanation}

LLMs generate tokens sequentially, one per round, until a stop criterion is met (e.g., reaching a stop token or the max token limit). Existing local explanation techniques, designed for single-output tasks like classification or regression~\cite{shickel2020sequential}, are not directly applicable to LLMs' token-by-token generation. To address this, we aggregate token importance scores generated by feature attribution techniques across each round of token generation.

\begin{figure}
    \centering
    \includegraphics[width=1\linewidth]{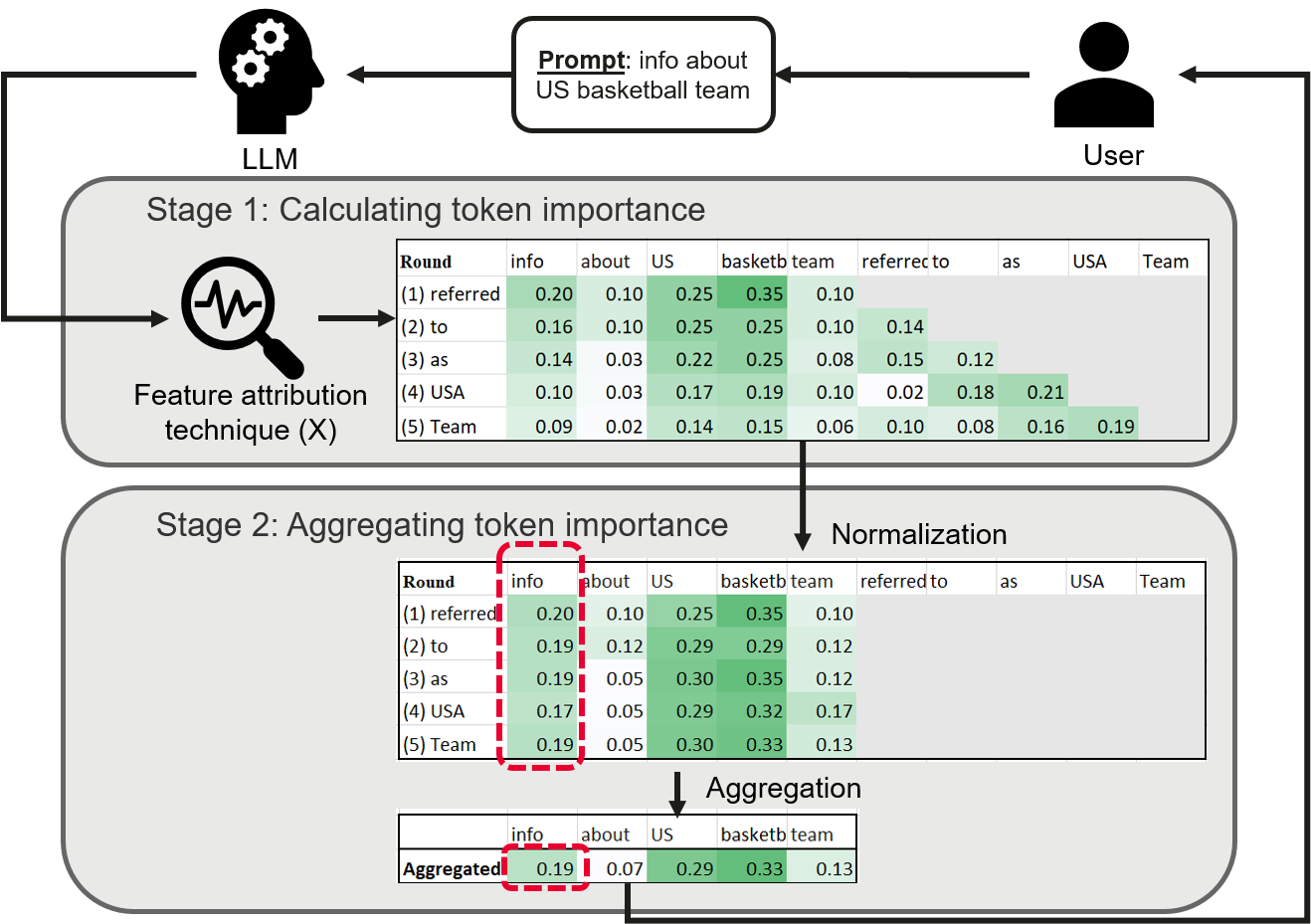}
    \caption{Workflow of aggregation-based approach. We show a running example with the prompt ``info about US basketball team''. In Stage 1, we calculate the importance score at token-level for each round of the generation. For instance, at the third round ``(3) as'', the importance scores for input tokens are ``info (0.14)'', ``about (0.03)'', ``US (0.17)'', ``basketball (0.25)'', ``team (0.08)'', ``referred (0.15)'', and ``as (0.18)''. In stage 2, we aggregate the importance scores for each token in the prompt across different rounds. For instance, the importance score for ``info'' is calculated based on the aggregation of ``info'' across all rounds highlighted in the red dashed box.}
    \label{fig:workflow}
\end{figure}

Our aggregation-based approach is divided into two stages. Stage 1 uses existing local explanation techniques to compute the importance score of prompt tokens across all rounds of token generation. In Stage 2, the token importance scores from all rounds are aggregated, representing the cumulative importance of prompt tokens across the entire sequence. These aggregated scores provide a comprehensive evaluation of the significance of each token of the prompt. Figure~\ref{fig:workflow} presents the workflow of the aggregation-based approach. 

\subsubsection{Stage 1: Calculating token importance}\label{sec:stage1}
In this stage, we first generate the importance of tokens in a prompt by leveraging the existing local explanation technique $X$.
Given a prompt $P$ = \{$p_1,p_2, ..., p_n$\},  LLM generates $Output$ = \{$o_1,o_2, ..., o_m$\}. For the $k$th round of token generation $o_k$, since the original prompt $P$ and generated tokens prior to $o_k$ (i.e., \{$o_1,o_2, ..., o_{k-1}$\} ) will be taken as the input for LLM, we will produce the importance for all those input tokens. Therefore, we calculate a matrix of a size $(n+m)\times m$, where the element $m_{ik}$ indicates the importance score generated by $X$ for the $i$th token in the concatenation of $p$ and $Ouput$ at $k$th round of generation.
For instance, given the $P$ = \{Info, about, US, basketball, team\}, LLM generates $Output$ = \{referred, to, as, Team, USA\}. The table in stage 1 in Figure~\ref{fig:workflow} shows the importance score matrix for this example. 

In this stage, we select to use a gradient-based technique, i.e., Integrated Gradients (IG)~\cite{sundararajan2017axiomatic} as our local explanation technique, because of its effectiveness in model explanation, time efficiency, and model-agnostic. Note that our technique is not limited to a specific explanation technique. Any effective, efficient, and model-agnostic explanation technique could be used in our framework.

\subsubsection{Stage2: Aggregating token importance}

As shown in Figure~\ref{fig:workflow}, each token $p_i$ in the prompt will be generated $m$ (i.e., the number of tokens in output) scores for it, and each presents their contribution to the output at the corresponding round of token generation. Since our goal is to provide a compact view of the prompt explanation, in this stage, we aggregate the importance scores for each token $p_i$ across rounds by using the following equation~\ref{eq:1}.

\begin{equation}\label{eq:1}
    imp(p_i) = \sum_{k=1}^M w_k *normalization(m_{ik})
\end{equation}

where $normalization(m_{ik})$ is the normalized importance score after eliminating the contribution from generated tokens and $w_K$ is the weight at the $k$th round. $M$ is the size of sampled rounds to aggregate. We will discuss how to select $M$ below.

We do the normalization since we focus on the prompt explanation, not on the generated tokens. For instance, for the second round of token generation in Figure~\ref{fig:workflow}, we calculate the importance score for six tokens info (0.16), about (0.10), US (0.25), basketball (0.25), team (0.25), and referred (0.14). After normalization, the importance scores for the prompt tokens are info (0.19), about (0.12), US (0.29), basketball (0.29), and team (0.12).

The time complexity of our aggregation-based approach is $O(M*X)$, where $M$ is the size of sampled rounds and $X$ represents the time complexity of the chosen local explanation technique. Once the local explanation technique is selected, its time complexity becomes fixed, making the overall complexity of our aggregation-based approach $O(M)$. We propose two weighting schemes and they have different sampling strategies. 
\begin{itemize}
    \item \textbf{Equal weighting} We assume all rounds of importance scores are equally important. Therefore the weight is $\frac{1}{M}$. We sample $M$ rounds evenly across the output sequence at intervals of $\frac{m}{M}$. This ensures we capture diverse samples across the entire sequence. For example, if the output length is 13 and $M$ is set to 5, we select rounds 1, 4, 7, 10, and 13 for aggregation, and the weight for each round is $\frac{1}{5}$.
    \item \textbf{Confidence weighting} We assume the contributions from each round are different. The output token with a higher confidence makes more contributions since the model are more confident in the output and in other words, the model probably understands the prompt better. Therefore, we use the confidence of each output token (i.e., probability) as the weight for the corresponding round. When sampling rounds to aggregate, we select the top $M$ rounds with the highest confidence scores. 
\end{itemize}

We refer to the aggregation-based approaches using these two schemes as Agg$_{Equ}$ and Agg$_{Conf}$, respectively. Unless stated otherwise, we set the value of $M$ to 5 for this study. The impact of the parameter $M$ on the effectiveness of our approach is discussed in Section~\ref{sec:rq2}.




\subsection{Perturbation-based Token-level Explanation}\label{sec:perb}

Previous studies show that perturbation is an effective black-box approach for model explanation~\cite{lundberg2017unified}. This approach involves systematically perturbing the input data and observing how these changes affect the model's output and helps identify which parts of the input are most influential in driving the model's predictions.

Although perturbation-based explanation could be directly used as a local explanation technique in Stage 1 in aggregation-based approach~\ref{sec:stage1}, it is highly time-consuming. Therefore, we adapt the perturbation-based approach to explain the prompt for LLMs directly. More specifically, given an LLM $L$, and prompt as $P$ = \{$p_1,p_2, ..., p_n$\}, we follow the steps below to calculate the importance score for each token in prompt $P$. 

\begin{itemize}
    \item \textbf{Step 1: Baseline Prediction}: The original prompt $P$ is passed through the LLM to obtain a baseline prediction. The output $Output_{Base}$ = $\{o_1, ..., o_i, ..., o_m\}$. The corresponding logits for each token in the vocabulary $V$ at each round (i.e., time step) are $\{logit(\{V_1, P), ..., logit(V_i, P), ..., logit(V_m, P)\}$ are recorded, where $logit(V_i, P)$ is the logit of each token in the vocabulary at $i$the round when querying the LLM with the original prompt $P$. We can access the logit of any token $x$ in the vocabulary $V$ at round $i$ using $logit(V_i, P)[x]$. 
    \item \textbf{Step 2: Prompt Perturbation}: One token in the original input is masked to create a perturbed version of the input. For instance, if the $j$th token is masked, the resultant prompt is denoted as $P_j$.  
    \item \textbf{Step 3: Impact Evaluation}: The perturbed input $P_j$ is fed back into the LLM. The corresponding logit for each token in the vocabulary at each time step are recorded as well, i.e., $\{logit(V_1, P_j), ..., logit(V_i, P_j), ...,  logit(V_g, P_j)\}$. We evaluate the impact of perturbation below.
    \item \textbf{Step 4: Repeating}: Step 1 to Step 3 is repeated for each token in the original prompt and their importance scores are calculated.
\end{itemize}

To evaluate the impact of perturbation (i.e., Impact Evaluation), we need to compare the output of the original prompt and that of perturbed prompts. We propose three approaches as below.

\noindent\textbf{Perb$_{Log}$:} Following prior studies~\cite{kokhlikyan2020captumunifiedgenericmodel}, we compare the logits of outputs between original prompt and perturbed prompts. We subtract the token logit of the perturbed prompt from the baseline logit to see the absolute change. More specifically, given the original prompt $P$ and its output $Output_{Base}$ = $\{o_1, ..., o_i, ..., o_m\}$, and the perturbed prompt $P_j$, we measure the differences between the output of $P$ and $P_j$ as the importance score for the $j$th token $p_j$ in the prompt $P$ using following equation:
\begin{equation}
\begin{split}
imp(p_j) = \frac{1}{m} \sum_{o_i \in Output_{Base}} & (LogSoftmax(logit(V_i, P)[o_i]) \\
& - LogSoftmax(logit(V_i, P_j)[o_i]))
\end{split}
\end{equation}

We use $LogSoftmax$ for its numerical stability and computational efficiency. Note that the output length of $P$ and $P_j$ may differ; if $P_j$'s output is shorter, we set $LogSoftmax(logit(V_i, P_j)[o_i])$ to zero when calculating importance scores. Additionally, some LLMs, like OpenAI's GPT-3.5, restrict access to the full vocabulary, limiting users to the top $K$ tokens at each time step. If token $o_i$ cannot be retrieved, we similarly set $LogSoftmax(logit(V_i, P_j)[o_i])$ to zero. Limited access to logits may reduce effectiveness, which we examine with varying $K$ values in Section~\ref{sec:rq2}.


Previous studies show that model inversion attacks can extract raw data using gradients or logits and possibly cause security issues and expose the information of LLMs~\cite{wang2024unique,ye2024defending}. Therefore, to address this issue, we also investigate two variants of perburbation-based approaches that do not require logit information. 

\noindent\textbf{Perb$_{Sim}$:} Instead of using the logit of each token to estimate the difference between the output of the original prompt $P$ and that of the perturbed prompt $P_j$, we calculate their semantic difference. To do so, we use SentenceBert~\cite{reimers2019sentence} to embed the output and measure their similarity using cosine similarity. Suppose the output of $P_j$ is $Output_j$, we calculate the importance score as follows:

\begin{equation}
    imp(p_j) = 1 - sim(embed(Output_j), embed(Output_{Base}))
\end{equation}
, where $sim()$ returns the cosine similarity between two outputs' embedding. A larger difference indicates the perturbed words are more important.

\noindent\textbf{Perb$_{Dis}$:} In this variant, we discretize logits of tokens to 0/1. If the output of the perturbed prompt $Output_j$ contains a token in $Output_{Base}$, we consider its logit is 1, otherwise is 0. Therefore, we measure the similarity of two outputs by counting their overlapped tokens following the below equation:

\begin{equation}
    imp(p_j) = 1 - \frac{|Output_{Base} \cap Output_j|}{|Output_j|}
\end{equation}

\subsection{User-defined granularity prompt explanation}\label{sec:userdeinfed}

We introduce token-level prompt explanation by computing the importance score of prompt tokens, representing the finest granularity. Aggregation could be especially helpful for complex prompts with multiple components (e.g., instruction, few-shot examples, etc.). Therefore, we offer the flexibility to extend prompt explanation to multiple granularities, such as word, sentence, or component by summing up the importance scores of corresponding tokens. With PromptExp, users have the freedom to define their desired granularity (e.g., [persona + instruction], [example1], [example2], etc.) and explore the importance of each defined component according to their preferences. This adaptability empowers users to conduct in-depth investigations into the significance of different prompt elements as they see fit. More specifically, suppose user define $l$ components $C_{defined}$ = \{$C_1, C_2, …, C_j,…, C_l\}$, where $C_j$ is the $j$th defined component, which is composed with $z$ tokens $\{p_1^j, p_2^j, …, p_z^j\}$. The importance score for $C_i$ can be computed as 
\begin{equation}
   imp(C_j) = \sum_{p_k^j \in C_j} imp(p_k^j) 
\end{equation}




\subsection{User Interface}

\begin{figure}
    \centering
    \includegraphics[width=1\linewidth]{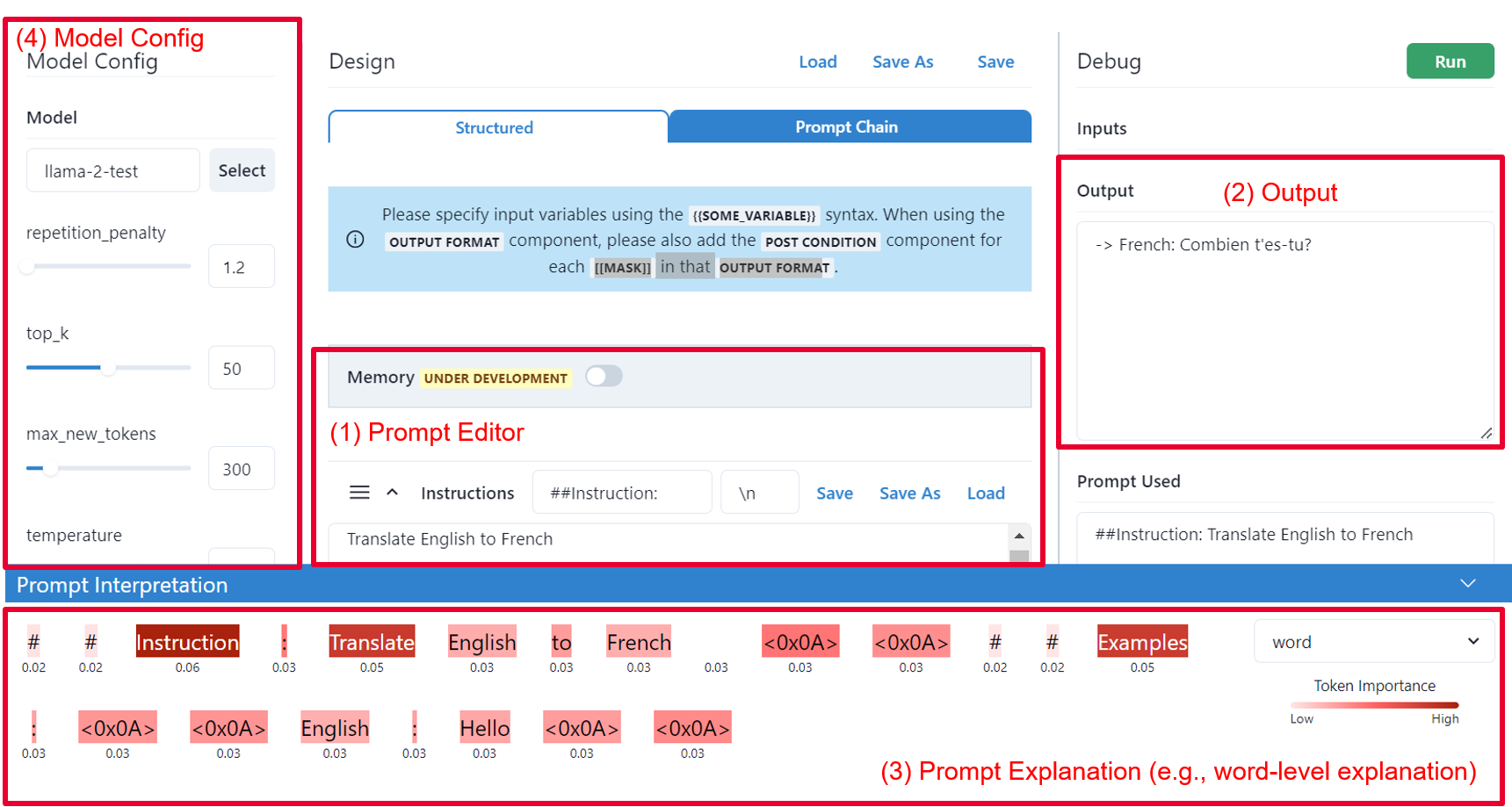}
    \caption{UI of PromptExp.}
    \label{fig:UI_wordLevel}
\end{figure}

We implement PromptExp as a browser-based UI, as shown in Figure~\ref{fig:UI_wordLevel}. PromptExp contains the following components:

\begin{itemize}
    \item \textbf{Prompt Editor:} Users can enter and edit their prompt in Prompt Editor, run the model, and generate the prompt explanation. Based on their observation, they can play around with their prompt and refine it iteratively based on the output and explanation. Note that users can define different components in this editor area.
    \item \textbf{Output:} The Output window displays the output of the model. 
    \item \textbf{Prompt Explanation:} The Prompt Explanation component returns the contributions of a prompt in different levels of granularity, e.g., word, sentence, and component with different colors. Darker colors indicate a more important contribution made by that part. PromptExp leverages feature attribution techniques (e.g.,  IG) to calculate the token-level importance score. To support the various needs of users and various complex levels of prompts, we extend it to different levels of granularity.
    \item \textbf{Model Config:} Users can select models and set model's parameters (e.g., Max tokens and Temperature). Note that explanation brings overhead. Therefore, users can choose to turn on and off the explanation function as they wish. 
\end{itemize}


\section{Experimental design}\label{sec:experimentalsetting}
We conducted a quantitative analysis to evaluate the effectiveness of PromptExp. We introduce the experimental design in the following sections.

\subsection{Research Question}\label{sec:rq}

We evaluate the various aspect of our framework by answering following research questions. 

\begin{itemize}
    \item RQ1: How effective is \ourTool?
    \item RQ2: What is the impact of different parameters on the effectiveness of \ourTool?
\end{itemize}


\subsection{Tasks and Evaluation}\label{sec:task}
Our goal is to evaluate whether the important components identified by our approach indeed have a significant impact on the output of a model. To do so, we design two tasks as discussed below.

\subsubsection{Task1: Sentiment classification}\label{sec:task1}
We select to conduct a case study on the sentiment classification task, since it is a widely selected case study for explainable AI studies~\cite{liu2023towards,wu2021polyjuice}. 
We design a prompt template to analyze the sentiment of a given sentence and our prompt is shown as follows:

\boxc{\textbf{Prompt} = [Query] [Instruction] \\
\textbf{Instruction} = ``analyze the sentiment of the previous sentence and respond only with POSITIVE or NEGATIVE. Your answer is''}

Our prompt has two components: Instruction and Query. The instruction part of the prompt instructs the Large Language Model (LLM) to analyze the sentiment of the sentence and the query is the sentence to be analyzed. 

In this task, we select the most widely used benchmark for sentiment analysis Stanford-Sentiment-Treebank (SST)~\cite{socher2013recursive} for our study. 

For each query, we compute the importance of each word, as well as each component (i.e., Instruction and Query).
Then, we perturb the most important tokens in the original query identified by our framework, and feed the perturbed query into the LLM, and observe if the outcome of the model changes. 
To measure the changes in models after the perturbation, we use \textbf{Flip rate (Flip)}~\cite{chrysostomou-aletras-2021-improving} as our measurement. The Flip can be calculated as the ratio between the number of times, the perturbation of the prompt resulted in a change in outcome from what the model originally predicted. 

To see if modifying the prompt based on our importance scores has a significant impact on the model outcome, we conducted an A/B test. More specifically, we first calculate Flips for the following two groups:
\begin{itemize}
    \item \textit{Treatment}: We begin by comparing the importance of the Instruction and Query components. Case 1: If the Instruction is identified as more important than the Query (Instruction $>$ Query), we select the top x\% most important words in the Instruction and replace them with random words, following prior studies~\cite{chrysostomou-aletras-2021-improving}. We then feed the perturbed prompt to the LLM to observe whether the predicted label flips. Case 2: If the Query is deemed more important than the Instruction (Instruction $<$ Query), we apply the same process to the \textit{top} x\% of the most important tokens in the Query and count the instances where the label flips. To address the possibility that the randomly selected words might have similar meanings to the original important words, we measure the semantic similarity between the perturbed and original prompts, ensuring that the similarity is below 0.7. We use SentenceBert~\cite{reimers2019sentence} for prompt embedding and cosine similarity for measuring semantic similarity. Finally, we calculate the Flip rate by summing the number of flips in both cases. For example, we have 200 cases, 180 cases are Instruction $>$ Query and 20 are Instruction $<$ Query. Among the 180 Instruction $>$ Query cases, we find 150 cases where the predicted label flips. Among the 20 Instruction $>$ Query cases, 10 cases' labels flip, then the Flip rate is $\frac{150+10}{200} = 80\%$.
    \item \textit{Control}: Similar to the Treatment group, we first identify which components (i.e., Query and Instruction) is more important, and then replace the \textit{bottom} x\% least important words with random words and examine if the predicted label flips or not. The Flip rate is calculated then. 
\end{itemize}

We noticed a small portion of the cases, where no sentiment labels (``NEGATIVE/POSITIVE'') are generated by LLMs when the original prompt is fed. We filtered out such cases since we assume LLMs cannot accomplish the sentiment classification task for them.

We then test the following Hypothesis: 
\begin{itemize}
    \item Hypothesis 1: Perturbing important words has a larger impact on the output than perturbing less important words. 
    We expect Flip of Treatment group $>$ Flip of Control group. 
\end{itemize}

If the difference between Treatment and Control is larger, we assume the explanation approach is more effective.


\subsubsection{Task2: Examining the relationship between Suffix importance and its impact}\label{sec:task2}

A specific suffix added to a prompt can significantly influence the model's output compared to the original prompt~\cite{wei2022chain,petrov2023prompting,hackmann2024word}. For example, appending the suffix ``think step by step'' instructs the model to generate reasoning in a more detailed and step-wise manner. Inspired by prior research~\cite{hackmann2024word}, we designed a task to investigate the relationship between the importance score of an added suffix and its impact on the output. Specifically, given a prompt $P$, we attach the suffix ``Give a short answer'' (i.e., $P_S$) and measure the word count difference between the outputs of the original prompt and the prompt with the suffix (i.e., delta length). We also compute the importance score of the suffix in the prompt $P_S$. If our method explains the prompt effectively, we expect a positive correlation between the importance score of the suffix and its impact, i.e., a shorter output, since the suffix instructs the LLM to provide a short response.

Similar to Task 1, we compute the correlation for two groups. For the treatment group ($Treatment$), we calculate Spearman’s correlation between the maximum importance score of tokens in the suffix and the delta length. In the control group ($Control$), we measure the correlation between the maximum importance score of tokens in the main part of the prompt (before the suffix) and the delta length. We then test the following hypothesis.

\begin{itemize}
    \item Hypothesis: The max importance score of tokens in suffix has higher correlation with delta length than the max importance score of tokens in other parts. We expect the correlation of $Treatment$ $>$ the correlation of $Control$. 
\end{itemize}

\subsection{Base large language model}
In our study, we employed two different LLMs as the base models: GPT-3.5~\cite{ChatGPT} and Llama-2~\cite{touvron2023llama}. More specifically, we use gpt-3.5-turbo\footnote{https://platform.openai.com/docs/models/gpt-3-5-turbo} and Llama-2-7b-chat-hf\footnote{https://huggingface.co/meta-llama/Llama-2-7b-chat-hf}, respectively. These models were chosen due to their representation of both commercial general-purpose LLMs and open-source LLMs, as well as their ranking as top performers in various tasks~\cite{leaderboard1,leaderboard2}. We set the temperature to 0 to reduce the randomness and use the default value the rest settings for both models. 



\subsection{Implementation details}
We begin by downloading the official checkpoints for LLaMa2 from HuggingFace and use the API provided by OpenAI to access gpt-3.5-turbo. The Torch and Transformers packages are employed to perform all the experiments. All experiments are done in Python 3.10. we use all-MiniLM-L6-v2~\cite{all-MiniLM-L6-v2} from sentence-transformers as the implementation of SentenceBERT in our experiment. All experiments were conducted on a machine equipped with a GPU of 24GB, a 24-core CPU, and 24 GB of RAM.

\subsection{Approaches for RQs}
\subsubsection{RQ1}
To evaluate the effectiveness of \ourTool, we evaluate both the aggregation-based and perturbation-based token-level explanation approaches on the two tasks using both LLMs (i.e., LLaMa-2 and GPT-3.5), and compare their performance. Note that GPT-3.5 is a closed-source model, we are not able to access the model to calculate the gradients for aggregation-based token-level explanation. Therefore, we focus on comparing perturbation-based token-level explanation approaches for GPT-3.5. As discussed in Section~\ref{sec:method}, GPT-3.5 is only allowed to access the top 20 tokens in the vocabulary at each time step. Therefore, if we cannot retrieve the specific token in the top 20 tokens returned by openAI's API, we set the logit for that token to zero. We ran the experiment 5 times and take the average as the results. 

\subsubsection{RQ2}
\ourTool has two parameters that can potentially impact its effectiveness, namely, $M$ and $K$, where $M$ determines the number of rounds to aggregate in the aggregation-based approaches and $K$ determines the size of vocabulary we can access calculating Perb$_{Log}$. In this RQ, we investigate 5, 10, and 30 for $M$ and 20, 1\% (320), 10\% (3,200), 50\% (16,000), and 100\% (32,000) of the vocabulary size for $K$. Intuitively, a smaller $M$ and $K$ typically means less information and results in a downgradation of effectiveness, while less computational cost. We wish to provide insights into the balance between effectiveness and efficiency.
For this RQ, we focused on the sentiment classification task using open-sourced LLM LLaMa2 since it provides logits for the full vocabulary and allows us to use IG.

\section{Results}\label{sec:results}

\subsection{RQ1 - Effectiveness}\label{sec:rq1}

\begin{table*}[]
\centering
\caption{The results of \ourTool with various proposed token-level explanation approaches. The Flip rate for Task 1 and Spearman's correlation for Task 2 are presented.}\label{tab:rq1}
\begin{tabular}{l|l|l|l|l|l|l|l|l|l}
\hline
  &   & \multicolumn{5}{c|}{\textbf{Llama-2}}         & \multicolumn{3}{c}{\textbf{GPT-3.5}}                                                        \\
                                        \hline
                                       
                                &    \textbf{Group}    & \textbf{Perb$_{Sim}$} & \textbf{Perb$_{Dis}$} & \textbf{Perb$_{Log}$} & \textbf{Agg$_{Conf}$} & \textbf{Agg$_{Equ}$} &\textbf{Perb$_{Sim}$} & \textbf{Perb$_{Dis}$} & \textbf{Perb$_{Log}$} \\
                                        \hline

\multirow{2}{*}{\textbf{Task 1}}  &  \textbf{\textit{Treatment}}                             & 0.68                        & 0.53                         & 0.6                       & 0.68        &      0.55          & 0.7                             & 0.55                      & 0.63                         \\

& \textbf{\textit{Control}}                               & 0.29                         & 0.38                         & 0.37                      & 0.32         & 0.36              & 0.08                          & 0.05                       & 0.09                         \\



\hline

\multirow{2}{*}{\textbf{Task 2}}  &  \textbf{\textit{Treatment}}                             & 0.33                         & 0.31                         & 0.21                        & 0.28         &      0.25          & 0.20                             & 0.21                      & -0.06                        \\

& \textbf{\textit{Control}}                          &             0.12            &           0.13             &          0.14             &    0.10      & 0.17              &    0.12                       &       0.13                 &          0.01                \\
\hline
\end{tabular}
\end{table*}

\textbf{In general, Perb$_{Sim}$ outperforms all other variants, including aggregation-based approaches. For aggregation-based approaches, Agg$_{Conf}$ outperforms Agg$_{Equ}$. } 
Table~\ref{tab:rq1} presents the results for 1 and Task 2. For both Llama-2 and GPT-3.5, the hypotheses are confirmed: the Flip rate of the treatment group is higher than that of the control group in both tasks. In general, an approach is considered better if it shows a higher Flip rate or correlation in the treatment group and a lower Flip rate or correlation in the control group. From this perspective, Perb$_{Sim}$ performs the best, while Perb$_{Dis}$ performs the worst. For instance, Perb$_{Sim}$ performs the best in Task 1 for both models and in Task 2 for Llama-2, whereas Perb$_{Dis}$ performs worst in both cases. This is expected since Perb$_{Dis}$ only measures token overlap, missing the logits information used in Perb$_{Log}$ and the semantic similarity used in Perb$_{Sim}$. Notably, Perb$_{Log}$ performs poorly on GPT-3.5, likely because only the top 20 logits are accessible, which reduces accuracy when assessing the impact of token perturbation.
When comparing aggregation-based approaches, Agg$_{Conf}$ consistently outperforms Agg$_{Equ}$, suggesting that weighting token importance by confidence across time steps is more effective than equal weighting.


\begin{figure*}
    \centering
    \subfigure[Agg$_{Conf}$]{\includegraphics[width=0.195\textwidth]{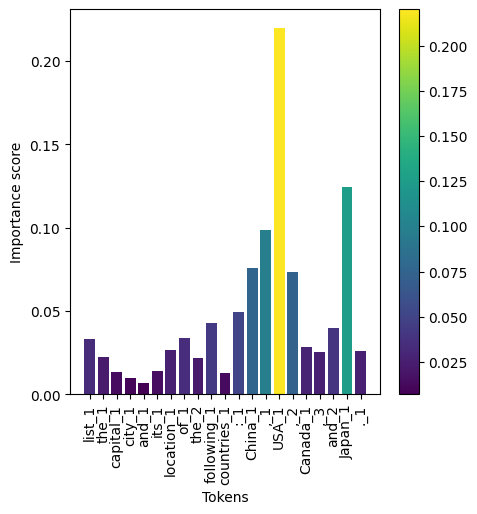}}
    \subfigure[Agg$_{Equ}$]
    {\includegraphics[width=0.197\textwidth]{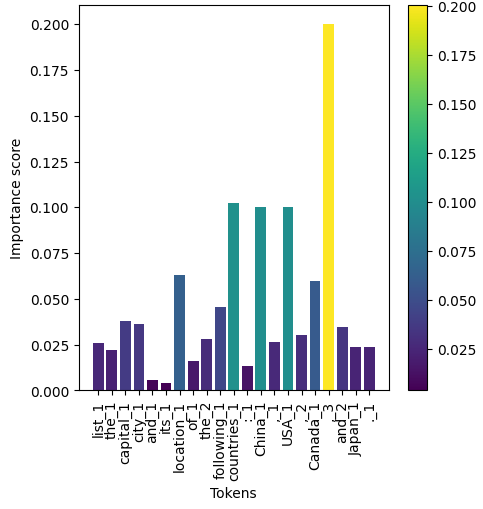}}
    \subfigure[Perb$_{Dis}$]
    {\includegraphics[width=0.19\textwidth]{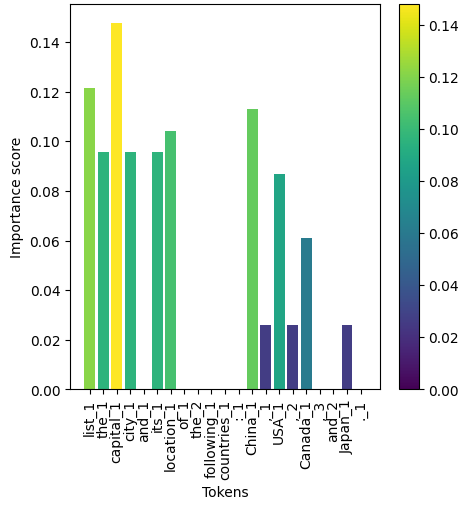}}
    \subfigure[Perb$_{Sim}$]{\includegraphics[width=0.189\textwidth]{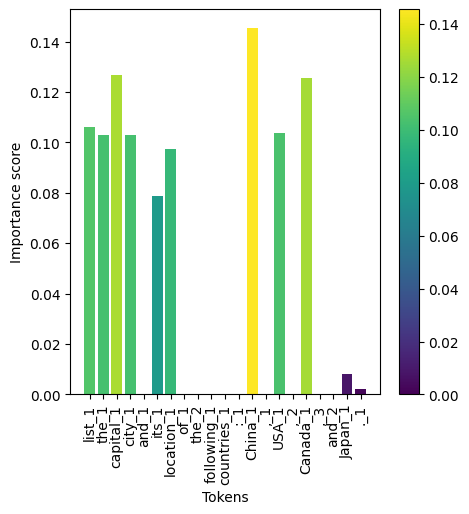}}
    \subfigure[Perb$_{Log}$]{\includegraphics[width=0.192\textwidth]{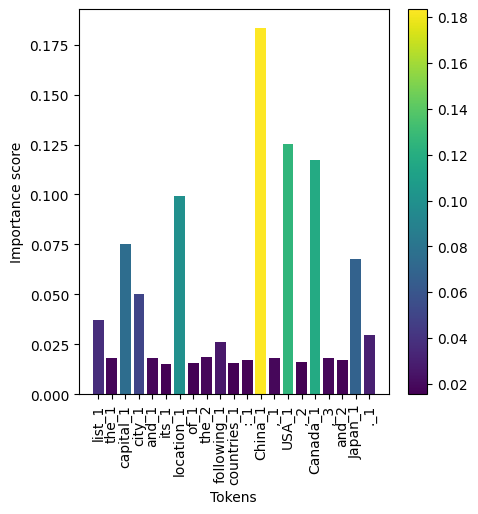}}
    \vspace{-0.1in}
    \caption{The visualization of a prompt ``list the capital city and its location of the following countries: China, USA, Canada, and Japan.'' with our proposed approaches. Note that Perb$_{Sim}$ and Perb$_{Dic}$ are more likely to have zero importance scores for certain tokens since they measure the similarity of output. If removing tokens does not change the output, the importance of that token is 0.}
    \label{fig:visulization}
\end{figure*}

Figure~\ref{fig:visulization} presents an example of the token-level explanation for a prompt with different token-level explanation approaches. Those approaches have different importance score distribution. The perturbation-based approach seems to provide explanations that align with the human sense. For instance, the most important tokens for this query in the human sense probably should be ``capital'', ``China'', ``USA'', and ``Canada''.  Those words all appear in the top list of perturbation-based approaches, while the important word ``capital'' is not in the top list of aggregation-based approaches.


\subsection{RQ2 - Impact of Parameters}\label{sec:rq2}

\begin{figure}
    \centering
    \subfigure{\includegraphics[width=0.25\textwidth]{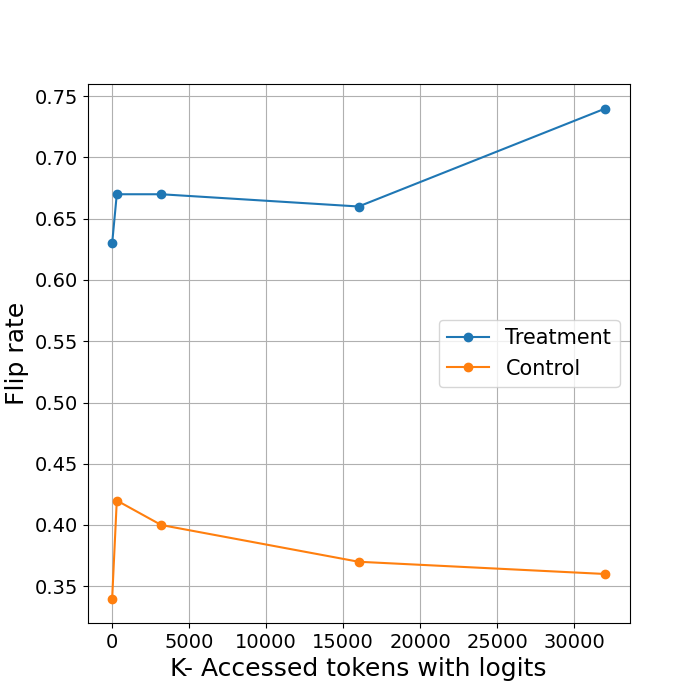}}
    \subfigure{\includegraphics[width=0.23\textwidth]{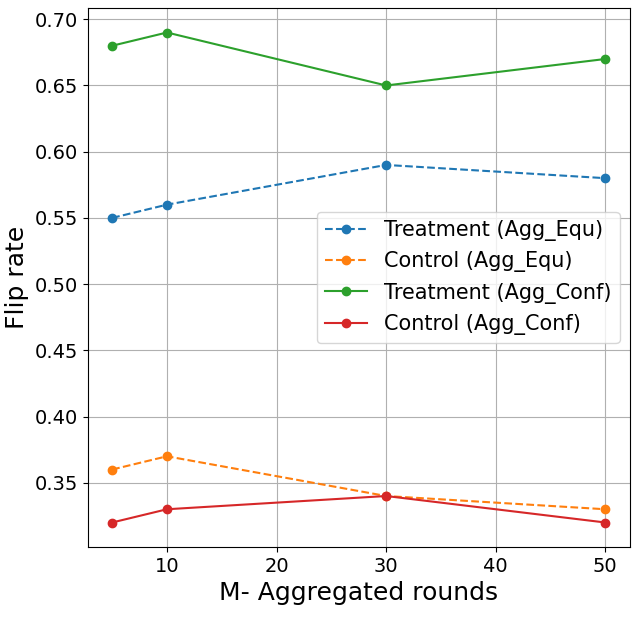}}
    \caption{Impact of $K$ on \ourTool using Perb$_{Log}$ (left), and impact of $M$ on \ourTool using Agg$_{Equ}$ and Agg$_{Conf}$ (right).}
    \label{fig:RQ2}
    \vspace{-0.2in}
\end{figure}

\textbf{Accessing logits of more tokens in the vocabulary improves the effectiveness of Perb$_{Log}$.} Figure~\ref{fig:RQ2} (left) shows the results for different values of $K$. While increasing $K$ does not result in a significant rise in the Flip rate for the treatment group, it does lead to a reduction in the Flip rate for the control group. This outcome is expected, as accessing the logits of more tokens provides a more accurate estimation of the impact of masking a token in Perb$_{Log}$.

\textbf{Increasing sampled rounds improves Agg$_{Equ}$ but does not enhance Agg$_{Conf}$.} Figure~\ref{fig:RQ2} (right) presents the results for different values of $M$. Increasing the number of sampled rounds has varying effects on Agg$_{Equ}$ and Agg$_{Conf}$. For Agg$_{Equ}$, the Flip for the treatment group rises, while the Flip for the control group decreases, which indicates an improvement in effectiveness. This is reasonable, as a larger $M$ increases the sample size, and incorporates more information. For Agg$_{Conf}$, the trend remains relatively stable, which suggests that sampling only the top rounds with the highest confidence is sufficient.

\section{User Study}\label{sec:userstudy}
\ourTool aims to help users better understand how prompts influence a model's behavior. More specifically, it helps users understand which components in the prompts have the most influence on the model's output. Therefore, we conduct a user study to evaluate the quality of the explanation generated by \ourTool and its usefulness in practice. 

\subsection{Participants} 
We invited 10 developers from the industry to participate in the user study. The developers have various levels of experience, e.g., entry-level (1-3 years), senior-level (4 - 6 years), and experienced (more than 6 years). The majority of them have more than 3 years of software development experience. We also ensured all those participants had experience using LLMs for various tasks.

\subsection{Tasks}

We designed three tasks for participants to assess the quality and usefulness of the explanation generated by \ourTool. More specifically, the first task is designed to assess the quality, and the second and third tasks are designed to assess the usefulness in practice.
\begin{itemize}
    \item \textbf{Task 1: Rating quality of explanation}  we ask participants to rate each explanation based on its reasonableness, using a scale of 1 to 5. A score of 1 signifies Strongly Disagree (the explanation is completely unreasonable or inaccurate), while a score of 5 indicates Strongly Agree (the explanation is entirely reasonable and accurate). For this evaluation, we sampled five prompts from five distinct tasks in the BigBench dataset~\cite{bigbench}, which comprises over 200 tasks and is widely used to evaluate large language models (LLMs). Each participant will be presented with five prompts, their outputs, and the corresponding explanations, and will be asked to rate the explanations. 
    \item \textbf{Task 2: Rating helpfulness for misclassification cases} Prior studies have demonstrated that interpretable explanations can uncover biases or misalignments between a model’s decision-making process and the intended task~\cite{ribeiro2016should,haffar2021explaining}. Model explanations offer valuable insights into the reasons behind misclassifications and help practitioners understand model behavior. In this task, we aim to evaluate whether the prompt explanations generated by \ourTool can aid in understanding misclassification cases in classification tasks. Specifically, we select four misclassification examples from a sentiment analysis task (Section~\ref{sec:task1}). As in Task 1, participants are provided with the prompt, the output label, and the corresponding explanation, and are asked to rate how helpful the explanation is in understanding the misclassification, using a scale of 1 to 5.  
    \item \textbf{Task 3: Prompt compression} The second task is to compress the prompt with the assistance of the explanation generated by \ourTool. Many LLMs have token limits and incur higher computational costs as the prompt length increases. Prompt compression reduces the length and complexity of inputs, therefore enhancing inference speed, reducing costs, and improving user experience.~\cite{jiang2023llmlingua,pan2024llmlingua,li2024500xcompressor}. Therefore, in this task, participants are asked to compress prompts by utilizing the explanations provided by \ourTool. Each participant is given two sample prompts from the Python Programming task in the BigBench dataset~\cite{bigbench} and is asked to perform prompt compression.
\end{itemize}

Note that we used the explanation generated by Perb$_{Sim}$ with Llama-2 for this user study, given its effectiveness as shown in Section~\ref{sec:results}.

\subsection{Procedure}
Each participant is given 20 minutes to complete the tasks and fill out a survey\footnote{https://forms.gle/kQcUDN3j4RumU5o47}. The user studies were conducted in a controlled environment. Prior to the study, we set up the PromptExp environment and provided participants with a tutorial on how to use it. During the study, the first author was present to assist participants with any questions. At the conclusion of the study, we also collected feedback on which aspects of the tool participants found most helpful and areas where improvements could be made.

\subsection{Results:}
\begin{figure}
    \centering
    \includegraphics[width=\linewidth]{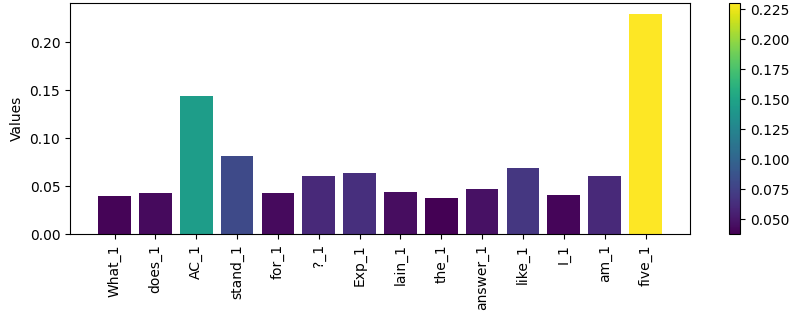}
    \vspace{-0.6cm}
    \caption{Word-level explanation for a prompt ``What does AC stand for? Explain the answer like I am five'' in Task 1 of the user study. The output is ``Years old. Sure, I'd be happy to explain what AC stands for in a way that a five-year-old can understand! AC stands for "Air Conditioning." ... ''.}
    \label{fig:userstudy_Q1}
    \vspace{-0.1in}
\end{figure}

\begin{figure}
    \centering
    \includegraphics[width=1\linewidth]{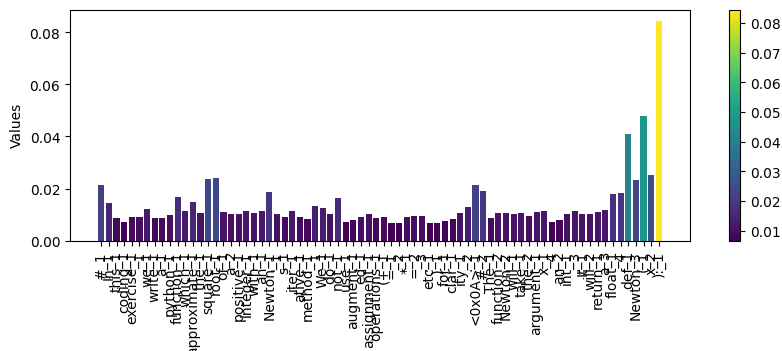}
    \vspace{-0.6cm}
    \caption{Word-level explanation for a prompt ``\# In this coding exercise, we write a python function which approximate the square root of a positive integer with an Newton's iterative method...'' in Task 1 of the user study. The output is ``:   y = x / 2 return (1 / (1 + (x / y)**2)) ...''.}\label{fig:userstudy_Q4}
    \vspace{-0.1in}
    
\end{figure}

\textbf{In terms of the quality of explanation, the average rating is 4.2. In 84\% (42/50) cases, participants strongly agree or agree that the explanation generated by \ourTool is reasonable and accurate.} For instance, Figure~\ref{fig:userstudy_Q1} presents the explanation for a prompt in Task 1. 8 out of 10 participants strongly agree that the prompt is reasonable and accurate. As the explanation shows, the most important words in the prompts are ``AC'', and ``five''. We also observed that the rating score of longer prompts is lower than shorter prompts. For instance, a prompt as shown in Figure~\ref{fig:userstudy_Q4} is constructed to solve a math problem and the prompt has 33 words. The rating for this prompt is 3.5. Although there are still 6/10 of participants believe the explanation is reasonable and accurate. However, two participants feel the explanation is not accurate. One possible reason is that when the prompt is long, too many tokens are labeled as important and confuse users, as one participant mentioned in their feedback `` when the prompt is long, as there are many tokens, the results of PromptExp may be difficult to interpret.''. This highlights the need for multi-granularity prompt explanation, typically when the prompt is long and complex. 

\begin{figure}
    \centering
    \includegraphics[width=1\linewidth]{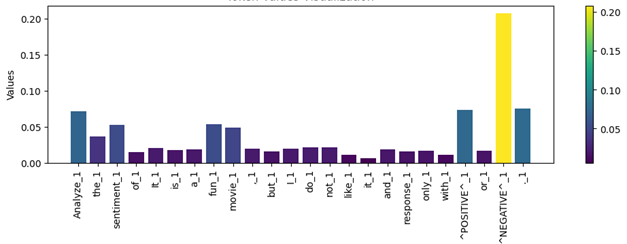}
    \vspace{-0.2in}
    \caption{Word-level explanation misclassification case where the prompt is ``Analyze the sentiment of It is a fun movie, but I do not like it and response only with \^{}POSITIVE\^{} or \^{}NEGATIVE\^{}:''. The output is ``\^{}POSITIVE\^{}'', while the true label is NEGATIVE.}
    \label{fig:userstudy_Q6}
    
\end{figure}

\textbf{For misclassification understanding (Task 2), the average rating is 3.82. In 65\% (27/40) of cases, participants strongly agreed or agreed that the explanations generated by \ourTool helped them understand the misclassification.} In this task, we conducted a user study to evaluate whether the explanations provided by \ourTool could assist participants in comprehending the model's behavior, particularly in instances of misclassification. The overall feedback is positive. In the majority of cases (27/40), participants found the explanations to be helpful. For example, Figure~\ref{fig:userstudy_Q6} illustrates an explanation for a sentiment misclassification case. In this case, 7 out of 10 participants agreed that the explanation was useful in understanding the model's behavior and the reason for the incorrect classification. The sentence, ``It is a fun movie, but I do not like it.'' was wrongly classified as ``POSITIVE". Upon closer inspection of the explanation, it becomes evident that the model overlooked the negative sentiment expressed in ``I do not like it'' and placed more emphasis on the phrase ``fun movie''.

\begin{figure}
    \centering
    \includegraphics[width=1\linewidth]{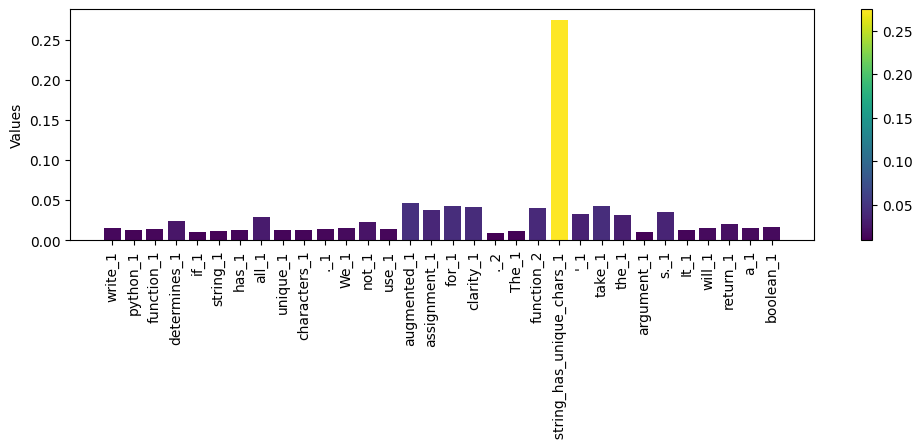}
    \vspace{-0.25in}
    \caption{Word-level explanation for the prompt ``write python function determines if string has all unique characters. We not use augmented assignment for clarity. The function 'string\_has\_unique\_chars' take the argument s. It will return a boolean.'' in prompt compression task.}
    \label{fig:userstudy_Q10}
    \vspace{-0.1in}
\end{figure}

\textbf{For prompt compression, the average rating is 4.2, In 80\% (16/20) of cases, participants strongly agreed or agreed that \ourTool can help them compress prompt.} In this task, we asked participants to use \ourTool to compress the given prompts. The feedback is positive and almost all participants found \ourTool is helpful in prompt compression. For instance, in the open-ended feedback, several participants mentioned that ``In my opinion, prompt compression is the most helpful feature.'', ``It is really helpful for compressing prompts. I can see which tokens/words have low importance and remove them from my prompts to save on inference costs.'', and ``I find the length reduction to be the most helpful.''. Here we illustrate how one participant leveraged \ourTool to compress prompts by removing non-important words. Figure~\ref{fig:userstudy_Q10} presents the explanation for the initial prompt. As we can see, the most important part is ``string\_has\_uniqu\_chars''. Therefore, the participant kept the most important words, and gradually removed less important phrases such as ``We not use augmented assignment for clarity''. After removing the less important words, the prompt was reduced to ``Python, The function `string\_has\_uniqu\_chars' take the arguments. It will return a boolean.''.

\subsection{Feedback for Further Improvement}

Despite the majority of the participants found \ourTool are useful in understanding prompt and model behavior, e.g., ``The ability to identify the most important part of the prompt in generating the current output is helpful in understanding why the model produces such results, which aids developers in refining and improving prompts.''. The participants also provide feedback to further improve \ourTool. We elaborate on several aspects below.

\noindent\textbf{\faThumbsUp Providing the option for users to define which part of the prompt should be the focus of the explanation.}
One valuable piece of feedback we received was: ``I think this tool should focus on specific tokens defined by the user instead of the entire set of tokens.'' Although \ourTool provides user-defined granularity for explanations, which allows multi-granularity options such as token-level, word-level, and component-level, the feedback suggests that users desire more control over which parts of the prompt are highlighted with importance scores, rather than applying this across the entire prompt. For instance, another participant noted: ``PromptExp gives too much importance to ``true/false" or ``positive/negative" key words from the question, which I think it's not the key focus for human's inference.'' This feedback indicates that, in some instances, the classification label in the prompt is not the primary focus for users. Therefore, a key feature to integrate into \ourTool is to provide more feasibility to users that allow them to define which part of the prompt they want to emphasize on top of multi-granularity.

\noindent\textbf{\faThumbsUp  Providing immediate results when perturbing a prompt.} In our perturbation-based approach, we perturb a prompt by masking each token in the prompt and assessing the resulting impact as a proxy for the token's importance. Immediate feedback for each perturbed prompt can offer valuable insights into the effect of each token in the prompt and present how the model responds to the change. As one participant noted in their feedback, ``It would be great if the immediate results from perturbing prompts could be displayed, as this might offer insights into how to improve the prompt." A similar concept has been explored in prior tools~\cite{promptTuning}, where multiple prompt variants are automatically generated and evaluated to assist users in refining and debugging prompts effectively.

\section{Discussion}

Both aggregation-based and perturbation-based approaches can be parallelized to reduce execution time. 

In aggregation-based approaches, local explanation methods are used to calculate importance scores for selected rounds of token generation. The time complexity for aggregation-based approaches is $O(M*X)$, where $M$ represents the number of rounds sampled for aggregation, and $X$ is the complexity of the local explanation technique used. Since the importance scores for these rounds are computed independently, and $M$ is relatively small (as shown in RQ2, increasing $M$ does not help to improve the effectiveness), the process can be parallelized to reduce the time complexity to $O(X)$. For example, in our study, with $M$ set to 5, the average explanation time and pure inference time (baseline) are shown in Figure~\ref{fig:time}. When generating a sequence of 30 tokens using Llama-2, pure inference time is 2.05 seconds, and Agg$_{Equ}$ and Agg$_{Conf}$ take 3.84 and 3.95 seconds, respectively. The actual overhead is the time of executing 5 times of IG, which is constant.

\subsection{Overhead and Performance optimization}

\begin{figure}
    \centering
    \subfigure{\includegraphics[width=0.24\textwidth]{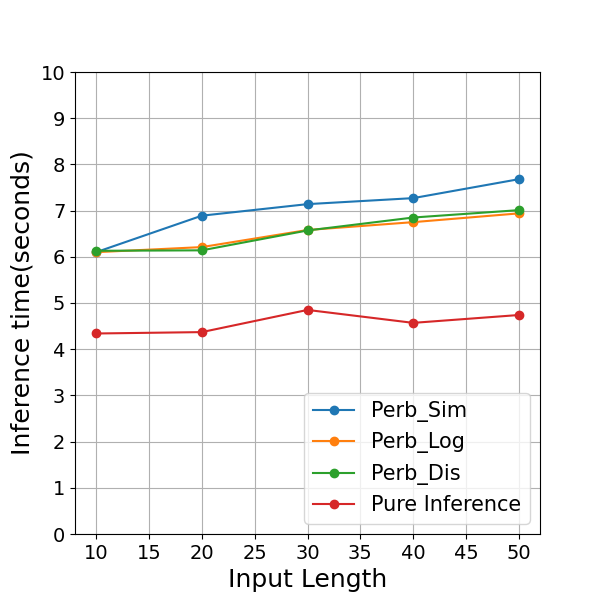}}
    \subfigure{\includegraphics[width=0.24\textwidth]{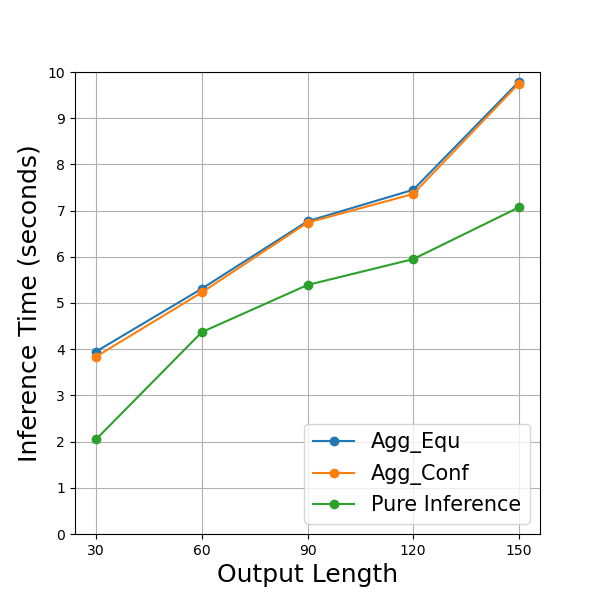}}
    \caption{Execution time of perturbation-based approaches against input length in token (left), and aggregation-based approaches against output length in token (right). Note that we include the pure inference time as a baseline. }
    \label{fig:time}
    \vspace{-0.2in}
\end{figure}

In perturbation-based approaches, each token in the original prompt must be perturbed, and its impact is assessed. The time complexity is $O(n*I)$, where $n$ is the length of the prompt, and $I$ represents the time complexity for evaluating the impact of perturbing a single token (i.e., the perturbed prompt). Since each perturbed prompt can be evaluated independently, this process can be parallelized by performing batch inference. Theoretically, this reduces the time complexity to $O(I)$, assuming no hardware limitations. However, server GPU memory constraints must be taken into account, as batch inference requires additional memory for each instance. For example, on Llama-2, adding an instance to a batch inference requires approximately 600 MB of GPU memory. With 24GB GPU, we can batch up to 15 instances after loading the Llama-2 model. Figure~\ref{fig:time} shows the average execution time of our perturbation-based approaches as input size increases, and keeps the output length fixed at 50 tokens. For instance, when the input size increases to 50, the average execution time for our perturbation approaches is around 7 seconds, compared to a pure inference time of 4.74 seconds, which is a reasonable overhead.

\subsection{Scope of different approaches}
We now discuss the applicability of different approaches. In our setting, we selected the gradient-based technique, Integrated Gradients (IG), as the local explanation method for the aggregation-based approach. IG is a white-box explanation technique that requires access to the entire neural network to compute gradients through backpropagation. As such, it is only suitable for cases where full access to the model is available, such as when working with locally deployed models. We chose IG in this study due to its efficiency. However, in theory, any local explanation method could be used, such as LIME~\cite{ribeiro2016should}.

In contrast, the perturbation-based approach is a black-box method that does not require access to the model's internals and only needs the output. We propose three perturbation-based methods: Perb$_{Log}$, Perb$_{Sim}$, and Perb$_{Dis}$, each requiring different levels of information. As introduced in Section~\ref{sec:perb}, Perb$_{Sim}$ and Perb$_{Dis}$ require the least information, i.e., only the output tokens, while Perb$_{Log}$ also requires access to logits. For scenarios with safety constraints where logits are not provided or only partially available, Perb$_{Sim}$ and Perb$_{Dis}$ are better options. In particular, Perb$_{Sim}$ has shown competitive performance on both LLMs, such as Llama-2 and GPT-3.5.

For practitioners prioritizing efficiency, Perb$_{Conf}$ is the optimal choice. It comes with less overhead compared to perturbation-based approaches while maintaining strong effectiveness, only slightly less effective than Perb$_{Sim}$.

\subsection{Threats to validity}\label{sec:threats}

\noindent\textbf{Internal Validity} 
Prompt engineering plays a crucial role in the performance of large language models (LLMs)~\cite{grabb2023impact}. Variations in prompts can lead to different outcomes, which may introduce inconsistencies in results. To address this, we standardized the prompt template across all experiments to ensure a fair comparison. In the user study, we recruited 10 developers as participants. While the sample size may not fully represent the broader population. We mitigated this threat by selecting developers with diverse levels of experience to capture a range of perspectives.

\noindent\textbf{External Validity}
relates to the generalizability of our findings. Since \ourTool is a framework the effectiveness of \ourTool also depends on the performance and capabilities of the LLMs. Our findings might not be generalized to other LLMs. To mitigate this threat, we evaluated \ourTool using two well-known and state-of-the-art LLMs, including both open-sourced and closed-sourced models.

\section{Related work}\label{sec:related}

\subsection{Local explanation for deep learning models}
Due to the ``black-box'' nature of the deep learning (DL) models and the importance of model explanation, an enormous of model explanation approaches have been developed for deep learning models~\cite{shickel2020sequential,zhao2024explainability}. One family of techniques called feature attribution-based explanation, which aims to measure the relevance of each input feature (e.g., words, phrases, text spans) to a model’s prediction~\cite{zhao2024explainability}, such as perturbation-based explanation~\cite{wu2020perturbed,li2015visualizing}, Gradient-based explanation~\cite{hechtlinger2016interpretation,sundararajan2017axiomatic,sikdar2021integrated}, surrogate models~\cite{ribeiro2016should,lundberg2017unified}, and decomposition-based methods~\cite{du2019attribution}. One family is called attention-based explanation, in which the attention mechanism is often viewed as a way to attend to the most relevant part of inputs~\cite{zhao2024explainability,jaunet2021visqa}. Intuitively, attention may capture meaningful correlations between intermediate states of input that can explain the model’s predictions. This family of approaches tries to explain models based on the attention weights or by analyzing the knowledge encoded in the attention~\cite{clark2019does,xu2015show}. Another family called example-based explanation~\cite{zhao2024explainability,koh2017understanding} intends to explain model behavior by illustrating how a model’s output changes with different input, such as adversarial examples~\cite{garg2020bae} and counterfactual explanations~\cite{cito2022counterfactual}. The last family is natural language explanation, in which a model is trained to generate explanation for input sequence~\cite{rajani2019explain,yordanov2021few}. In this study, we leverage existing local explanation techniques (i.e., IG and perturbation-based) in our framework to explain prompts for LLMs. In theory, any above-mentioned local explanation can be leveraged by our framework. However, computation complexity is a key point that practitioners should consider. 

\subsection{Explanation for LLMs}
One recent direction for LLM explanation is to use LLM's self-explain capability~\cite{singh2024rethinking,bhattacharjee2023llms,camburu2018snli}. For instance, chain-of-thought (CoT) prompting exemplifies this
approach, where an LLM is prompted to explain its reasoning step-by-step before arriving at an answer~\cite{wei2022chain}. Camburu et al. ~\cite{camburu2018snli} proposed an approach to use explanation-annotated data to train models, so that the trained models can output explanations along with answers. Despite their potential benefits, natural language explanations remain susceptible to hallucination or inaccuracies~\cite{chen2023models,ye2022unreliability}. In addition, such approaches do not directly explain the prompt and tell practitioners which components in the prompts are important. To address this, in this study, we propose a framework \ourTool to directly indicate which components make significant contributions to the output.


\section{Conclusion}\label{sec:conclusion}

We presented \ourTool, a framework that explains prompts by aggregating token-level explanation and offers the flexibility to analyze of LLM outputs at different levels of granularity. To implement the token-level explanation for LLMs, we propose two approaches aggregation-based and perturbation-based approaches, which cover both the white-box and black-box explanations. Our case studies show that in general, the perturbation-based approach using semantic similarity to measure the impact of perturbation performs the best among all proposed approaches. Additionally, a user study demonstrates the accuracy of the generated explanation and practical usefulness of \ourTool. Finally, we provide practical insights on selecting suitable token-level explanation techniques within \ourTool in practice.

\clearpage
\balance

\bibliographystyle{IEEEtran}
\bibliography{0-main_IEEE}

\end{document}